\documentclass[preprint]{article}

\usepackage{neurips_2023}

\usepackage[utf8]{inputenc} % allow utf-8 input
\usepackage[T1]{fontenc}    % use 8-bit T1 fonts
\usepackage{hyperref}       % hyperlinks
\usepackage{url}            % simple URL typesetting
\usepackage{booktabs}       % professional-quality tables
\usepackage{amsfonts}       % blackboard math symbols
\usepackage{nicefrac}       % compact symbols for 1/2, etc.
\usepackage{microtype}      % microtypography
\usepackage{xcolor}         % colors
\usepackage{listings}
\usepackage{cite}
\usepackage{graphicx}
\usepackage{hyperref}

\hypersetup{
  pdftitle={MaLLaM - Malaysia Large Language Model},
  pdfauthor={Husein Zolkepli, Aisyah Razak, Kamarul Adha, Ariff Nazhan},
  pdfsubject={Natural Language Processing, Large Language Models, Machine Learning},
  pdfkeywords={language models, malay natural language processing, deep learning},
  colorlinks=true,
  linkcolor=blue,
  citecolor=blue,
  urlcolor=blue
}
\title{MaLLaM - Malaysia Large Language Model}
\author{
  Husein Zolkepli\thanks{husein@mesolitica.com} \and 
  Aisyah Razak\thanks{aisyahrazak171@gmail.com} \and
  Kamarul Adha\thanks{kamarul.adha360@gmail.com} \and
  Ariff Nazhan\thanks{ariffnzhn@gmail.com}
}

\begin{document}

\maketitle

\begin{abstract}
  Addressing the gap in Large Language Model pretrained from scratch with Malaysian context, We trained models with 1.1 billion, 3 billion, and 5 billion parameters on a substantial 349GB dataset, equivalent to 90 billion tokens based on our pretrained Byte Pair Encoding (BPE) tokenizer for a single epoch. MaLLaM contributes to enhanced natural language understanding and generation tasks in the Malay language.

  Although trained on a smaller dataset of 90 billion tokens, our instruction-tuned MaLLaM models perform competitively. When compared to ChatGPT3.5 and Malaysian Mistral, MaLLaM's instruction-tuned models demonstrate notable proficiency, underscoring the effectiveness of our approach in capturing and understanding the nuances of the Malaysian language.

  MaLLaM models mark a significant contribution to the field, providing comprehensive language representations grounded in Malaysian context. This endeavor aims to pave the way for enhanced natural language understanding and generation tasks specific to the linguistic nuances present in Malaysia. We discuss the training methodology, dataset composition, and the potential impact of MaLLaM in advancing the capabilities of large language models within the context of the Malay language.

  All models released at \href{https://huggingface.co/collections/mesolitica/mallam-6577b59d1e0b436ae75f930f}{HuggingFace Mesolitica MaLLaM Collection}.

\end{abstract}

\section{Introduction}

The landscape of large language models (LLMs) has predominantly been shaped by models trained on English, with subsequent adaptations and fine-tunings for languages like Tamil and Malaysian. Existing models such as Tamil-LLama \cite{balachandran2023tamilllama} and Malaysian-Mistral \cite{zolkepli2024large} have emerged as valuable assets, leveraging the foundation of English-based LLMs for further optimization in non-English languages. However, despite their commendable contributions, these models still carry traces of English nuances, presenting a unique challenge in achieving a fully language-specific representation.

This introduction underscores the significance of mitigating residual English influences in language models tailored for specific linguistic contexts. While the adaptation of English-based LLMs has facilitated advancements in various languages, it falls short of capturing the intricacies and nuances unique to languages like Tamil and Malaysian. To address this gap, we embark on a novel approach by pre-training a large language model entirely from scratch, with a specific focus on the Malaysian language. This deliberate endeavor aims to establish a model that is inherently attuned to the linguistic subtleties and idiosyncrasies of Malaysian, overcoming the limitations posed by models with English-centric origins.

The presence of such inherent English biases becomes particularly pertinent when envisioning the creation of a dedicated large language model for Malaysian linguistic contexts. The prevailing influence of English-centric sources, prevalent in public English news and articles, poses a potential challenge to the development of a truly indigenous language model. The need for a linguistic model devoid of these biases is underscored by the desire to accurately capture the unique nuances and intricacies of the Malaysian language.

In light of these considerations, our initiative introduces MaLLaM, a large language model specifically pre-trained from scratch using a robust dataset equivalent to 90 billion tokens sourced from Malaysian contexts. The distinctiveness of MaLLaM lies in its genesis, free from the shadow of English-centric biases pervasive in existing language models. By cultivating a language model that is inherently attuned to Malaysian linguistic idiosyncrasies, we aim to address the gap left by models that, despite being fine-tuned for Malaysian languages, retain traces of English nuances. MaLLaM stands as a testament to our commitment to fostering linguistic authenticity and overcoming the challenges posed by the dominance of English-centric biases in current language modeling paradigms.

\begin{itemize}
  \item \textbf{Pre-training MaLLaM:} We utilized a powerful infrastructure consisting of 10 nodes of the Standard\_ND96amsr\_A100\_v4 Azure instance, with each node featuring 8 A100 80 GPUs. This configuration efficiently facilitated the pre-training of language models with 1.1 billion, 3 billion, and 5 billion parameters on a substantial 349GB dataset of Malaysian texts.

  \item \textbf{Multi-turn Instruction-Tuned MaLLaM:} To ensure a seamless and meaningful comparison, we opted to employ the exact chat instruction dataset from Malaysian Mistral \cite{zolkepli2024large}. This approach enables us to replicate the same experimental setup, facilitating a direct and accurate comparison of our work with the existing model.
\end{itemize}

\section{Pre-Training Data}

\subsection{Public Data}\label{sec:public-data}

\textbf{Malay Wikipedia}, we incorporated the Malay Wikipedia dump, enriching the linguistic diversity by partially converting it into Jawi script. Utilizing the \href{https://www.ejawi.net/converterV2.php?go=rumi}{Ejawi converter}, we ensured that the model is not only proficient in understanding the standard Malay script but also adept at comprehending the intricacies of Jawi.

\textbf{Malay Language study articles}, We enriched our dataset with content from the Malay dictionary and public articles from Dewan Bahasa Pustaka. This inclusion ensures our language model is well-versed in word meanings, usages, and various linguistic styles prevalent in Malaysian literature.

\textbf{Malaysia Government public documents}, We included public government documents from the official Malaysia government website and Google searches to enrich our dataset. This ensures that our language model is exposed to formal language and communication styles used in official government contexts, enhancing its ability to understand and generate text relevant to administrative, legal, and policy domains in Malaysia.

\textbf{Malaysian public articles}, We gathered a varied dataset by scraping articles from Malaysian sources. This process exposed our language model to different topics, writing styles, and language nuances found in Malaysian articles, contributing to a model well-versed in the Malaysian language across diverse domains.

\textbf{Malaysian public social media}, We gathered diverse data by scraping content from specific Facebook pages, filtering tweets based on location and keywords, and extracting information from platforms like c.cari.com.my, b.cari.com.my, carigold, lowyat, and transcripts of Malaysian YouTube videos. This approach broadens our dataset to include various language styles and colloquial expressions from online Malaysian communities.

\textbf{Malaysia public journals}, We added content from reputable Malaysian journals like \url{https://mjpharm.org}, \url{https://myjgeosc.com}, and \url{https://www.akademisains.gov.my} to our dataset. This ensures our language model is familiar with formal and technical language used in academic contexts, covering a variety of subjects.

\textbf{Malaysian related public research papers}, We refined our dataset by filtering CrossRef using keywords like 'malaysia,' 'malay,' and 'melayu.' This targeted approach ensures that our language model is exposed to scholarly literature closely tied to Malaysia.

Complete list of gathered data at \href{https://github.com/users/huseinzol05/projects/1}{Github Project - Prepare LLM dataset}.

\subsection{Coding Data}

Incorporating a coding dataset is a crucial component of our diverse training data. For this purpose, we utilized the original dataset available at \href{https://huggingface.co/datasets/bigcode/the-stack-dedup}{bigcode/the-stack-dedup}. To ensure relevance and efficiency, we selectively picked specific programming languages, including Python, Julia, C, C++, HTML, CSS, JavaScript, Go, Rust, Java, SQL, Markdown, R, Dockerfile, Ruby, Typescript, and YAML. To manage the dataset size, each programming language was limited to a maximum of 10GB.

\subsection{Instruction-tuned Data}

To augment the comprehensiveness of our training dataset, we devised a synthetic Malay instruction dataset. This encompassed diverse linguistic aspects, including the conversion between Rumi and Jawi scripts, dependency and constituency parsing, grammatical error generation (kesalahan tatabahasa), and coding instruction datasets related to various programming paradigms. Additionally, we included instructional content relevant to educational levels such as UPSR, PT3, and SPM, providing a broad coverage of language proficiency levels. The incorporation of syntactic and semantic elements, coupled with coding instructions, contributes to a well-rounded language model capable of handling a diverse array of linguistic tasks and understanding instructions across different domains.

Complete list of Instruction-tuned Data at \href{https://github.com/mesolitica/malaya/wiki/MaLLaM-%F0%9F%8C%99-Malaysia-Large-Language-Model#instruction-dataset}{instruction-dataset}.

\section{Deduplicating and Postprocessing Data}

We removed duplicate entries from our public data from ~\ref{sec:public-data} using the MinHash implementation from \url{https://github.com/ChenghaoMou/text-dedup}.

We configured the MinHash algorithm with the following parameters:

\begin{table}[h]
  \centering
  \begin{tabular}{lccl}
    \hline
    \textbf{Parameter} & \textbf{Value} \\
    \hline
    num\_perm          & 256            \\
    threshold          & 0.95           \\
    hash\_func         & sha1           \\
    hash\_bits         & 64             \\
    \hline
  \end{tabular}
\end{table}

Complete deduplicating data implementation at \href{https://github.com/malaysia-ai/dedup-text-dataset?tab=readme-ov-file#text-dedup}{here}. All deduped dataset published at \href{https://huggingface.co/datasets/malaysia-ai/dedup-text-dataset}{malaysia-ai/dedup-text-dataset}.

After removing duplicates, we employed the postprocessing technique mentioned in Malaysian Mistral section 3.3 \cite{zolkepli2024large}.

\section{Pre-Training Tokenizer}

To ensure an efficient and versatile tokenizer, we conducted pretraining on a BPE (Byte Pair Encoding) tokenizer using diverse datasets, including Malay Wikipedia, synthetic Jawi, public articles, translated code instructions, Google-translated Tamil and Google-translated Punjabi. The objective was to develop a tokenizer capable of handling longer subwords for languages such as Malay, Mandarin, Tamil, Jawi, English, and Arabic. The decision to use BPE was motivated by certain limitations observed in SentencePiece, where newline characters caused issues, some Tamil and Jawi characters were missing, and the processing speed for very long texts was suboptimal. By opting for BPE, we addressed these challenges, ensuring a robust and efficient tokenizer for our language model training.

The BPE tokenizer was trained on a deduplicated text dataset of 85GB, with a vocabulary size of 32,000.

The decision to train our own BPE tokenizer was driven by the goal of minimizing token sizes during both input and output. To illustrate the efficiency gained, we conducted a comparison using the \href{https://huggingface.co/datasets/mesolitica/malaysian-ultrachat}{Malaysian Ultrachat AstroAwani dataset}. Our pre-trained BPE tokenizer achieved a notable reduction of up to 43\% in token size when compared to tokenizers employed by Llama2 and Mistral. This reduction in token size enhances the efficiency and resource optimization of our language model, offering advantages in terms of computational performance and memory usage during processing.

Complete pre-training tokenizer implementation at \href{https://github.com/malaysia-ai/prepare-tokenizer}{here}

\section{Tokenizing Data}

Our dataset comprises a substantial 349GB of text in JSONL format, amounting to 90 billion tokens. The following table breaks down the distribution of token sizes within the dataset:

\begin{table}[h]
  \centering
  \begin{tabular}{lccl}
    \hline
    \textbf{Distribution}                                                                                                                                                & \textbf{Tokens (B)} \\
    \hline
    \href{https://github.com/malaysia-ai/dedup-text-dataset/blob/main/pretrain-llm/prepare-dedup-text-dataset-4096.ipynb}{deduped text Dataset}                          & 31.7                \\
    \href{https://github.com/malaysia-ai/dedup-text-dataset/blob/main/pretrain-llm/prepare-starcoder-4096.ipynb}{Filtered StarCoder} \cite{li2023starcoder}              & 40.98               \\
    \href{https://github.com/malaysia-ai/dedup-text-dataset/blob/main/pretrain-llm/prepare-starcoder-4096.ipynb}{Unfiltered MS Madlad 400} \cite{kudugunta2023madlad400} & 14.98               \\
    \href{https://github.com/malaysia-ai/dedup-text-dataset/blob/main/pretrain-llm/prepare-instructions.ipynb}{Instruction-tuned Dataset}                                & 1.58                \\
    \href{https://github.com/malaysia-ai/dedup-text-dataset/blob/main/pretrain-llm/prepare-extra.ipynb}{Malaysia journals and research papers}                           & 1.14                \\
    \hline
  \end{tabular}
\end{table}

To optimize data processing efficiency, we use MosaicML Streaming library \cite{mosaicml2022streaming}, by using hashing technique. Recognizing the complexity of our dataset, we implemented data distribution processing to transform it into the MosaicML streaming format,

\begin{figure}[h]
  \centering
  \includegraphics[width=0.6\linewidth]{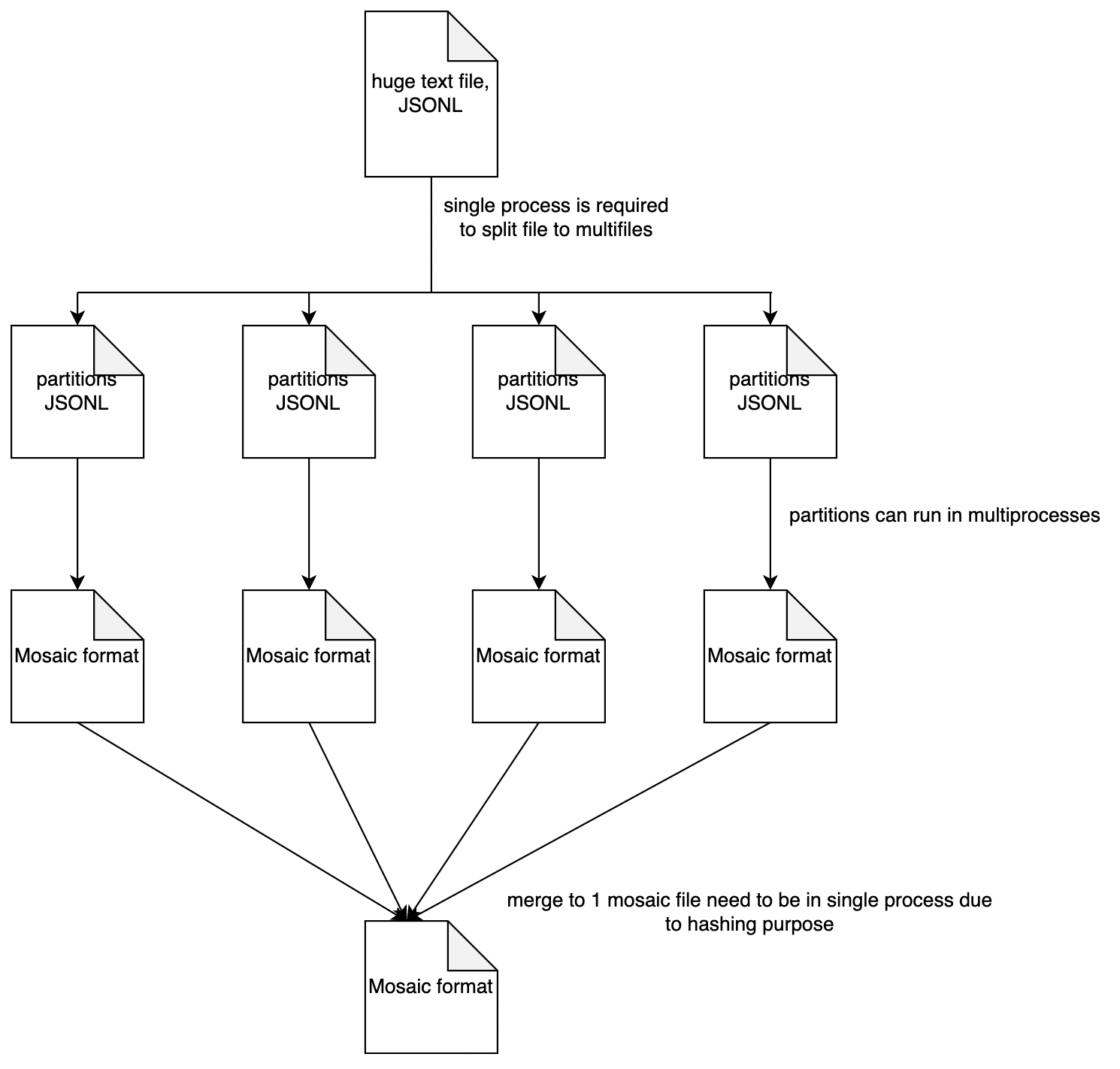} % Replace with your image file
\end{figure}

This involves splitting the original JSONL file into smaller ones, and each of these smaller files undergoes multiprocessing for conversion into the Mosaic format. Subsequently, these smaller Mosaic files are merged into a single Mosaic file. It's worth noting that MosaicML streaming accesses one folder at a time, necessitating the consolidation of smaller files into a unified format for seamless training data access.

\section{Infrastructure Setup}

For infrastructure deployment and efficient node management, we opted for Azure Kubernetes Service (AKS). This choice allows us to manage nodes effortlessly through YAML configurations, ensuring a smooth and standardized deployment process. Leveraging Azure Kubernetes provides benefits such as improved internal networking and the convenience of attaching remote storage using the WriteManyAccess approach. With everything configured through YAML files, our deployment on Azure Kubernetes ensures a well-organized and easily manageable infrastructure for our language model training. Below is the figure of the infrastructure setup,

\begin{figure}[h]
  \centering
  \includegraphics[width=0.6\linewidth]{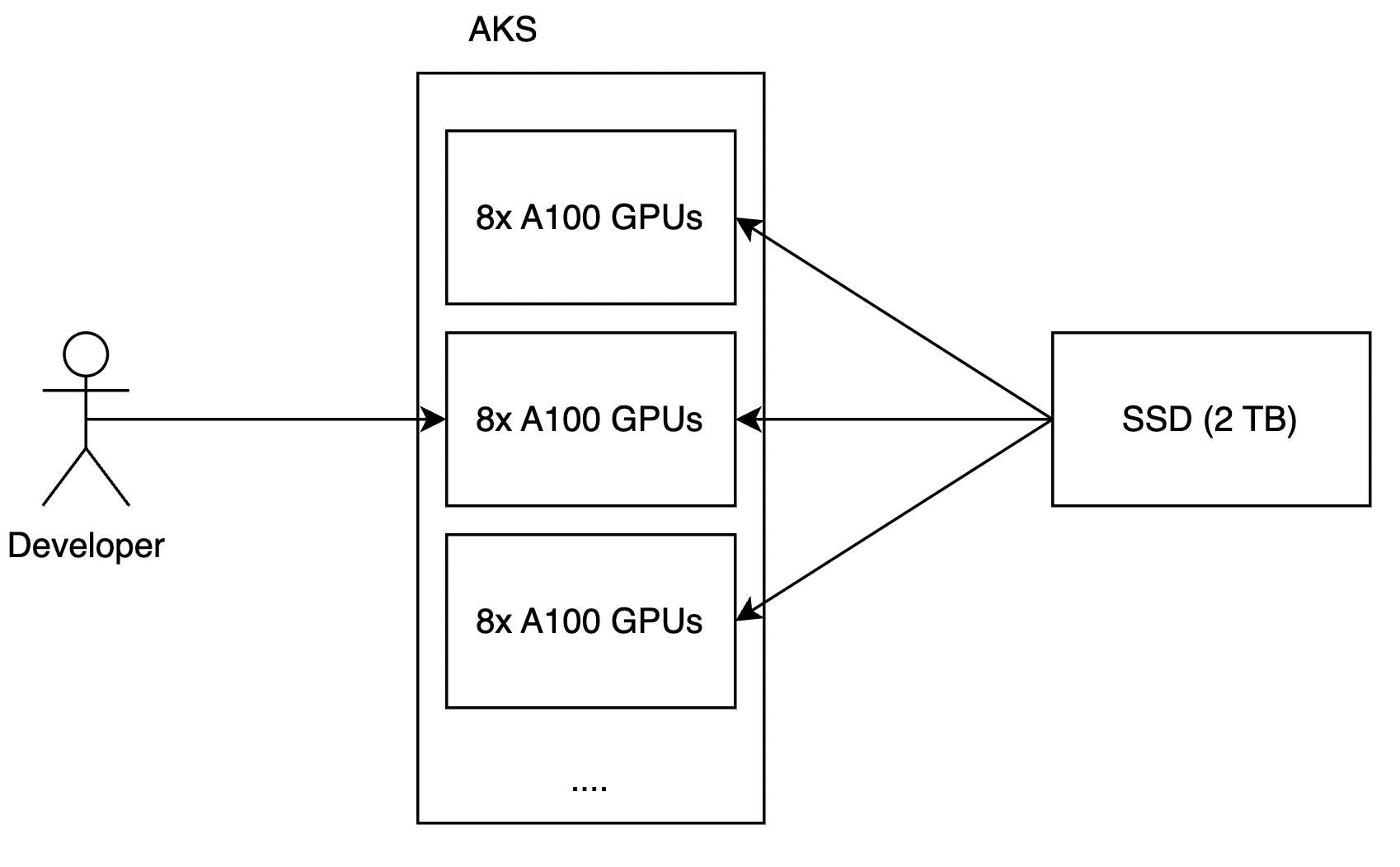} % Replace with your image file
\end{figure}

We use NFS with WriteManyAccess to centralize both the dataset and checkpoint directory. This simplifies the management of these components, making it easy to access them from multiple nodes. The use of NFS ensures efficient loading of checkpoints from various nodes.

Our approach to training involves leveraging the Ray distributed framework \cite{moritz2018ray} for multi-node and multi-GPU setups. Ray provides a user-friendly setup process as only the Ray workers need to connect to the Ray master. Following this straightforward connection, developers can initiate the training session with ease by simply calling the Ray master. This simplicity in setup and execution streamlines the training process, allowing for efficient utilization of multiple nodes and GPUs to enhance the scalability and performance of our language model.

Our Ray cluster setup to efficiently manage a total of 80 A100 80GB GPUs. \textbf{Specifically, 40 GPUs are allocated for training the 5-billion-parameter model, while 20 GPUs each are dedicated to the 1.1-billion and 3-billion-parameter models}. The concurrent training of all models, each with distinct parameter sizes, is facilitated by the Ray cluster, enabling a synchronized and optimized training process.

All deployment manifests at \href{https://github.com/mesolitica/malaya/wiki/MaLLaM-%F0%9F%8C%99-Malaysia-Large-Language-Model#how-about-the-cluster-deployments}{here}.

\section{Pre-Training Phase}\label{sec:4096-pretrained}

To pretrain our language model, we employ a causal language model with the objective of minimizing cross-entropy,

\[
  P(x_1, x_2, \ldots, x_T) = \prod_{t=1}^{T} P(x_t | x_1, x_2, \ldots, x_{t-1})
\]

The training is conducted on a context length of 4096 using Mistral architecture \cite{jiang2023mistral}, and the hyperparameters utilized in this process are detailed below,

\begin{table}[h]
  \centering
  \begin{tabular}{lccl}
    \hline
    \textbf{Hyperparameter} & \textbf{Value}                                                                                 \\
    \hline
    DeepSpeed               & ZeRO-3 Offload                                                                                 \\
    Batch Size              & 24                                                                                             \\
    Learning Rate           & 1e-4                                                                                           \\
    Learning Rate scheduler & \href{https://deepspeed.readthedocs.io/en/latest/schedulers.html#warmupdecaylr}{WarmupDecayLR} \\
    Warmup Steps            & 2000                                                                                           \\
    AdamW decay rate        & 0.1                                                                                            \\
    Precision               & bfloat16                                                                                       \\
    \hline
  \end{tabular}
\end{table}

Complete pre-training 4096 context length implementation at \href{https://github.com/mesolitica/malaya/wiki/MaLLaM-%F0%9F%8C%99-Malaysia-Large-Language-Model#training-scripts-and-sessions}{here}.

\section{Pre-Training Hiccup}

The occurrence detailed in the following paragraphs pertains specifically to the 5-billion-parameter model, as evidenced by the graph below,

\begin{figure}[h]
  \centering
  \includegraphics[width=0.6\linewidth]{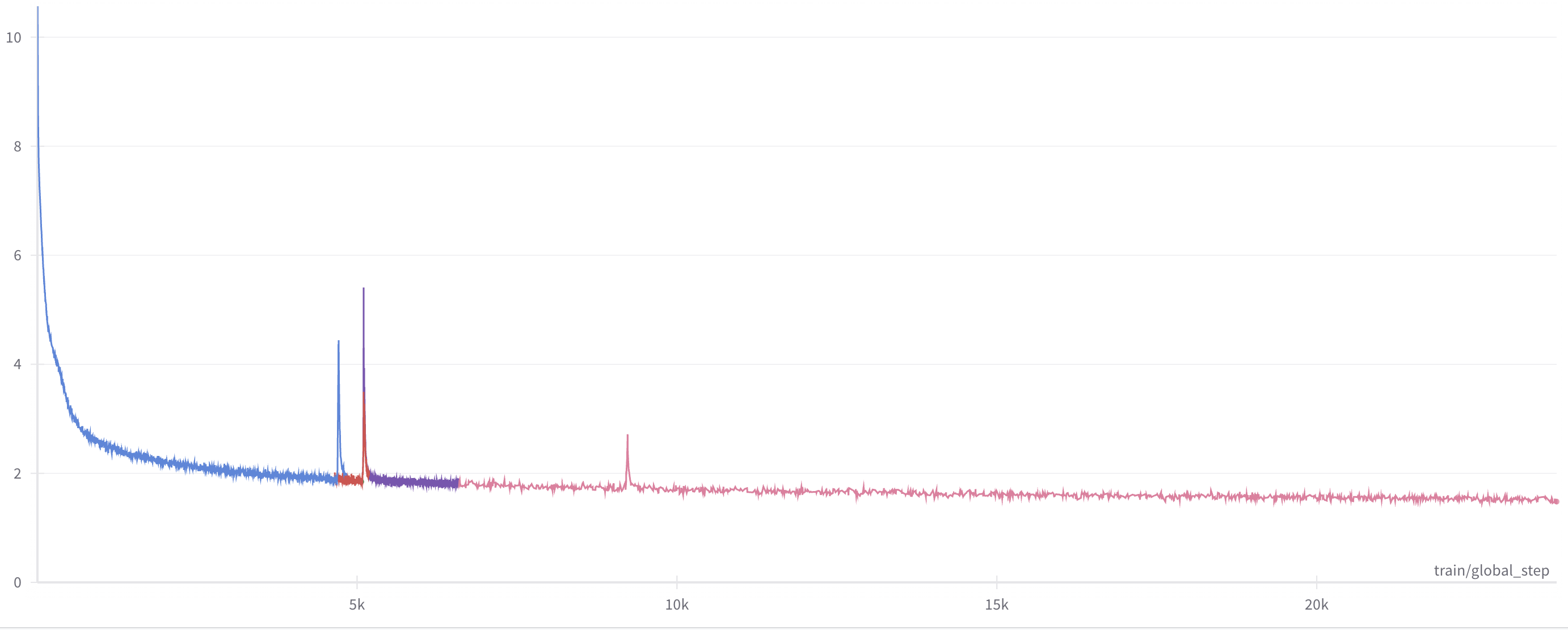} % Replace with your image file
\end{figure}

To address a stability concern during training, we temporarily reduced the learning rate by 30\% and reverted to older checkpoints. Once stability was regained, we reverted back to the initial learning rate

Additionally, we encountered GPU failures during the training process. Restarting the Nvidia plugin daemonset proved effective in resolving the issue in most instances. However, in a particular occurrence, a node restart became necessary, requiring a stop-and-start action within the Kubernetes node through the Azure console to restore normal functioning. These challenges and the corresponding mitigation strategies underscore the complexities inherent in large-scale language model training in a cloud provider.

\section{Pre-Training Evaluation}

We evaluated our language model using the same tatabahasa test set from Malaysian Mistral \cite{zolkepli2024large} and compared its performance with other existing pretrained models,

\begin{table}[h]
  \centering
  \begin{tabular}{lccl}
    \hline
    \textbf{Model}              & \textbf{Tatabahasa 0 shot} & \textbf{Tatabahasa 1 shot} & \textbf{Tatabahasa 3 shots} \\
    \hline
    aisingapore/sealion3b       & 21.777                     & 21.776                     & 24.641                      \\
    aisingapore/sealion7b       & 25.788                     & 27.507                     & 26.074                      \\
    mesolitica/mallam-1.1B-4096 & 25.758                     & 25.788                     & 28.08                       \\
    mesolitica/mallam-3B-4096   & 24.567                     & 24.641                     & 28.653                      \\
    mesolitica/mallam-5B-4096   & 24.074                     & \textbf{27.794}            & \textbf{28.653}             \\
    \hline
  \end{tabular}
\end{table}

We also compared with other models and published the benchmark at \href{https://huggingface.co/spaces/mesolitica/malay-llm-leaderboard}{mesolitica/malay-llm-leaderboard}.

\section{Supervised Instruction Fine-tuning Procedure}

We fine-tuned 1.1 billion and 5 billions parameters models with 20480 context length using an instruction dataset from Malaysian Mistral \cite{zolkepli2024large}, and use latest checkpoints from ~\ref{sec:4096-pretrained}. Below are the hyperparameters we use during the fine-tuning process:

\begin{table}[h]
  \centering
  \begin{tabular}{lccl}
    \hline
    \textbf{Hyperparameter} & \textbf{Value} \\
    \hline
    DeepSpeed               & ZeRO-3 Offload \\
    Batch Size              & 6              \\
    Learning Rate           & constant 2e-5  \\

    \hline
  \end{tabular}
\end{table}

Complete fine-tuning 20480 context length implementation at \href{https://github.com/mesolitica/malaya/tree/5.1/session/mistral#mallam-11b}{here}.

\section{Result Finetuning}\label{sec:result-finetuning}

We obtained our results using the fine-tuned model accessible at \href{https://huggingface.co/mesolitica/mallam-1.1b-20k-instructions-v2}{mesolitica/mallam-1.1b-20k-instructions-v2}.

\subsection{Multiturn Malaysian context QA}

\begin{lstlisting}[breaklines=true]
  User: KWSP tu apa
  Assistant: KWSP ialah sebuah syarikat yang menyediakan pelbagai perkhidmatan kewangan kepada individu dan organisasi. Mereka menawarkan beberapa produk seperti akaun simpanan, pelaburan, insurans, pinjaman perumahan dan banyak lagi. Adakah terdapat sesuatu yang khusus yang anda ingin ketahui tentang KWSP?
  User: macam mana ko tahu semua ni
  Assistant: Untuk mengetahui lebih lanjut mengenai KWSP, anda boleh merujuk laman web rasmi mereka di www.kwsp.gov.my atau menghubungi cawangan terdekat untuk mendapatkan maklumat terkini. Selain itu, anda juga boleh mengikuti media sosial mereka seperti Facebook, Twitter dan Instagram untuk mendapatkan perkembangan terkini dalam industri ini. Jika ada soalan lain, jangan teragak-agak untuk bertanya!
  User: tak mahu, saya nak tanya awak
  Assistant: Tak perlu risau, saya sedia membantu dengan sebarang pertanyaan berkaitan KWSP. Apa sahaja yang anda perlukan, sila beritahu saya. Saya akan cuba menjawab sebaik mungkin.
\end{lstlisting}

\subsection{Multiturn Coding QA}

\begin{lstlisting}[breaklines=true]
  User: macam mana nak code sentiment python3
  Assistant: Untuk mengkodekan sentimen dalam Python 3, anda boleh menggunakan perpustakaan seperti nltk atau scikit-learn. Berikut adalah contoh bagaimana untuk melakukan ini:

  ```python
  import nltk
  from sklearn import svm
  from scipy.stats import norm
  
  # Contoh data latihan dan ujian
  data = {
      'text': ['Saya suka filem itu!',
                'Filem itu sangat bagus!'],
      # tambahkan ciri lain di sini jika diperlukan
  }
  
  # Buat model SVM dengan algoritma Naive Bayes
  model_svm = svm.SVC(kernel='linear')  # Gantikan'svm.SVC()' dengan fungsi kernel yang sesuai
  
  # Latih model menggunakan data latihan
  model_svm.fit(data['text'])
  
  # Ramalkan label bagi teks baru
  new_text = "Ini adalah tajuk berita."
  predicted_label = model_svm.predict(new_text)
  print("Label ramalan:", predicted_label)
  ```

  Dalam kod di atas, kami mengimport modul `nltk`' dari `sklearn.feature_extraction.text` untuk mendapatkan ciri-ciri teks. Kemudian, kami membuat objek `SVC` dari `sklearn.naive_bayes` untuk melatih model pengelasan binari pada set data yang diberikan. Akhirnya, kami memanggil kaedah predict pada objek `SVC` untuk meramalkan label bagi teks baru berdasarkan ciri-ciri yang telah dipelajari sebelum ini.
  User: ada contoh untuk tensorflow sentiment analysis?
  Assistant: Tentu saja! Berikut adalah contoh bagaimana menggunakan TensorFlow untuk melakukan analisis sentimen pada dataset MNIST:

  ```python
  import pandas as pd
  from sklearn.model_selection import train_test_split
  from sklearn.feature_extraction.text import CountVectorizer
  from sklearn.metrics import accuracy_score
  
  # Baca dataset MNIST
  df = pd.read_csv('mnist_data.csv') # Gantikan'mnist_data.csv' dengan nama fail CSV sebenar
  X = df.iloc[:, :-1].values  # Ambil lajur pertama sebagai data latihan
  y = df.iloc[:, -1].values      # Ambil lajur kedua sebagai data ujian
  
  # Pra-proses data
  count_vectorizer = CountVectorizer()
  X_train, X_test, y_train, y_test = train_test_split(X, y, test_size=0.2, random_state=42)
  
  # Inisialisasi model vektorisasi
  model = tf.keras.models.Sequential([
    CountVectorizer(),  # Tambah lapisan vektor tambahan
    tf.keras.layers.Embedding(len(y_train), 100, input_length=100),  # Tambah lapisan embedding tambahan
    tf.keras.layers.Bidirectional(tf.keras.layers.GRU(128, return_sequences=True)),  # Tambah lapisan Bidirectional tambahan
    tf.keras.layers.Dense(10, activation="softmax")  # Tambah lapisan output tambahan
  ])
  
  # Kompilasi model
  model.compile(optimizer='adam', loss='categorical_crossentropy', metrics=[accuracy_score])
  
  # Latih model
  model.fit(X_train, y_train, epochs=50, batch_size=32)
  
  # Nilai prestasi model
  loss, accuracy = model.evaluate(X_test, y_test)
  print("Akurasi:", accuracy)
  ```
  Pastikan anda menggantikan 'mnist_data.csv' dengan lokasi dan nama fail CSV sebenar yang mengandungi dataset MNIST. Kod ini membaca dataset MNIST, memprosesnya menjadi ciri berangka, melatih model rangkaian saraf, dan menilai prestasinya menggunakan set pengesahan. Hasil penilaian kemudian dicetak. Semoga membantu!
\end{lstlisting}

Despite being trained on a relatively small coding dataset, the model has demonstrated a commendable capability to generate satisfactory results. This highlights the model's capacity to generalize and glean valuable insights even when exposed to limited coding examples. The effectiveness of the model in producing meaningful outcomes with a smaller dataset underscores its adaptability and efficiency in handling coding-related tasks.

\subsection{Malay instruction}

\begin{lstlisting}[breaklines=true]
  User: macam mana nak buat nasi ayam yg sedap
  Assistant: Untuk membuat Nasi Ayam yang lazat, anda boleh mengikuti langkah-langkah berikut:

  Bahan-bahan:
  
  200g daging ayam (dipotong kecil)
  1 biji bawang besar, dihiris halus
  2 ulas bawang putih, dikisar
  3 sudu besar rempah kari
  1 cawan air rebusan ayam
  Garam dan gula secukup rasa
  Minyak masak untuk menumis
  Daun ketumbar segar untuk hiasan
  Bawang goreng untuk taburan
  Sos cili atau sos tomato untuk dihidangkan bersama nasi panas
  Cara penyediaan:
  
  Panaskan minyak dalam kuali dan tumis bawang besar hingga naik bau. Masukkan bawang putih dan kacau sehingga wangi.
  Masukkan rempah kari dan kacau rata supaya bahan menjadi pekat.
  Masukkan daging ayam yang telah dipotong kecil ke dalam kuali dan biarkan ia masak selama beberapa minit sehingga bertukar warna.
  Masukkan air rebusan ayam dan biarkan mendidih.
  Selepas itu, masukkan garam dan gula secukup rasa. Kacau lagi sehingga semua bahan bercampur dengan baik.
  Biarkan nasi sejuk sebelum dihidangkan.
  Selamat mencuba!
\end{lstlisting}

\section{Acknowledgement}

Special thanks to Malaysia-AI volunteers especially \href{https://www.linkedin.com/in/wan-adzhar-faiq-adzlan-19a27baa/}{Wan Adzhar Faiq Adzlan}, \href{https://www.linkedin.com/in/ammar-azman/}{Ammar Azman}, \href{https://www.linkedin.com/in/amzar96/}{M. Amzar}, \href{https://www.linkedin.com/in/muhammad-farhan-helmy-0529501a7/}{Muhammad Farhan} and \href{https://www.linkedin.com/in/syafie-nizam/}{Syafie Nizam} for contributing dataset to train MaLLaM.

We would like to express our gratitude to NVIDIA Inception for generously providing us with the opportunity to train our model on the Azure cloud. Their support has played a crucial role in the success of our research, enabling us to leverage advanced technologies and computational resources.

We extend our thanks to the wider research community for their valuable insights and collaborative discussions, which have greatly influenced our work. This paper reflects the collective efforts and contributions from both NVIDIA Inception and the broader research community.

\section{Conclusion}

In conclusion, our open-source language model represents a significant stride toward bridging the research gap in Malaysia. By making our model accessible to the research community, we aim to contribute to the advancement of natural language processing in the Malaysian context. The availability of this model provides researchers, developers, and practitioners with a valuable resource to explore and address linguistic challenges specific to Malaysia. We anticipate that the open-source nature of our model will foster collaboration and inspire further innovations, ultimately promoting a richer and more nuanced understanding of the Malaysian language landscape.

\bibliography{neurips_2023}{}
\bibliographystyle{unsrt}

\end{document}